\documentclass[10pt,twocolumn,letterpaper]{article}

\usepackage[pagenumbers]{iccv} 

\usepackage[export]{adjustbox}

\definecolor{iccvblue}{rgb}{0.21,0.49,0.74}
\usepackage[pagebackref,breaklinks,colorlinks,allcolors=iccvblue]{hyperref}

\newcommand{\ourcell}{\cellcolor[rgb]{1,0.808,0.576}}

\def\paperID{*****} 
\def\confName{ICCV}
\def\confYear{2025}

\title{AQUATIC-Diff: Additive Quantization for \\ Truly Tiny Compressed Diffusion Models}

\author{Adil Hasan, Thomas Peyrin\\
Nanyang Technological University, Singapore \\
{\tt\small \{s240139@e.ntu.edu.sg, thomas.peyrin@ntu.edu.sg}
}

\begin{document}
\maketitle

\begin{abstract}
Significant investments have been made towards the commodification of diffusion models for generation of diverse media. Their mass-market adoption is however still hobbled by the intense hardware resource requirements of diffusion model inference. Model quantization strategies tailored specifically towards diffusion models have been useful in easing this burden, yet have generally explored the Uniform Scalar Quantization (USQ) family of quantization methods. In contrast, Vector Quantization (VQ) methods, which operate on groups of multiple related weights as the basic unit of compression, have seen substantial success in Large Language Model (LLM) quantization. In this work, we apply codebook-based additive vector quantization to the problem of diffusion model compression.  Our resulting approach achieves a new Pareto frontier for the extremely low-bit weight quantization on the standard class-conditional benchmark of LDM-4 on ImageNet at 20 inference time steps. Notably, we report sFID 1.92 points lower than the full-precision model at W4A8 and the best-reported results for FID, sFID and ISC at W2A8. We are also able to demonstrate FLOPs savings on arbitrary hardware via an efficient inference kernel, as opposed to savings resulting from small integer operations which may lack broad hardware support.
\end{abstract}

\section{Introduction}
\label{sec:intro}

\textbf{Diffusion Models} (DM) \citep{ho2020denoising, Dhariwal2021DiffusionMB, Rombach2021HighResolutionIS} have become the dominant architecture for many tasks. The intense hardware resources involved in the iterative process of diffusion model inference have however proven a serious impediment. Knowledge distillation \citep{salimans2022progressive, meng2023distillation} and efficient sampling strategies \citep{song2020denoising, liupseudo, lu2022dpm} have seen success in reducing the number of model forward passes (denoising time steps) required for high-quality inference -- down to as little as twenty steps. However, with the latest open-source diffusion models such as Stable Diffusion 3 \citep{Esser2024ScalingRF} boasting of 8 billion parameters in total, the GPU VRAM and FLOPs requirements of a single forward pass are now a serious hindrance towards diffusion model inference on mass-market consumer hardware.

\begin{figure}[htbp]
\setlength{\tabcolsep}{1pt}
\begin{adjustbox}{max width=\linewidth} %
\begin{tabular}{l l l l}
\toprule
\huge Full Precision & & \huge 2 bits & \\
\midrule
\includegraphics[width=0.25\textwidth]{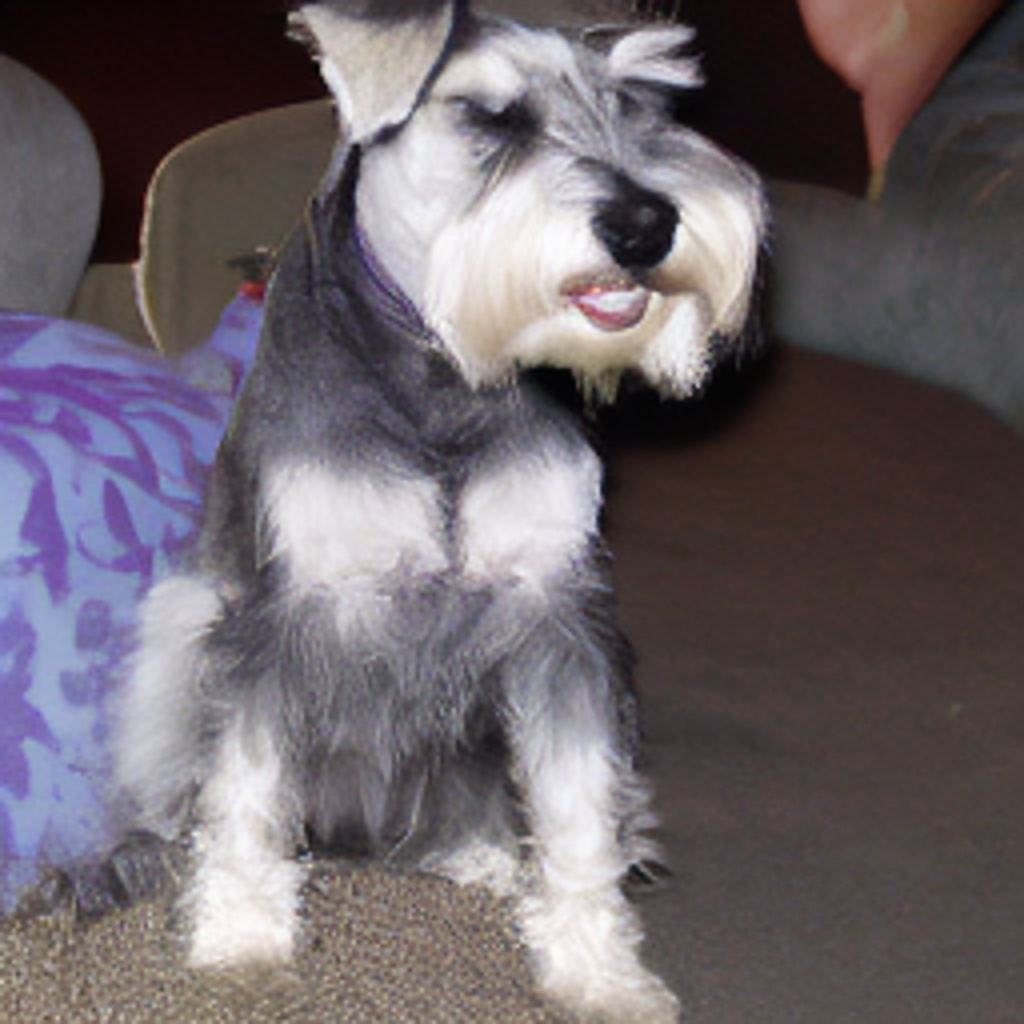} &%
    \includegraphics[width=0.25\textwidth]{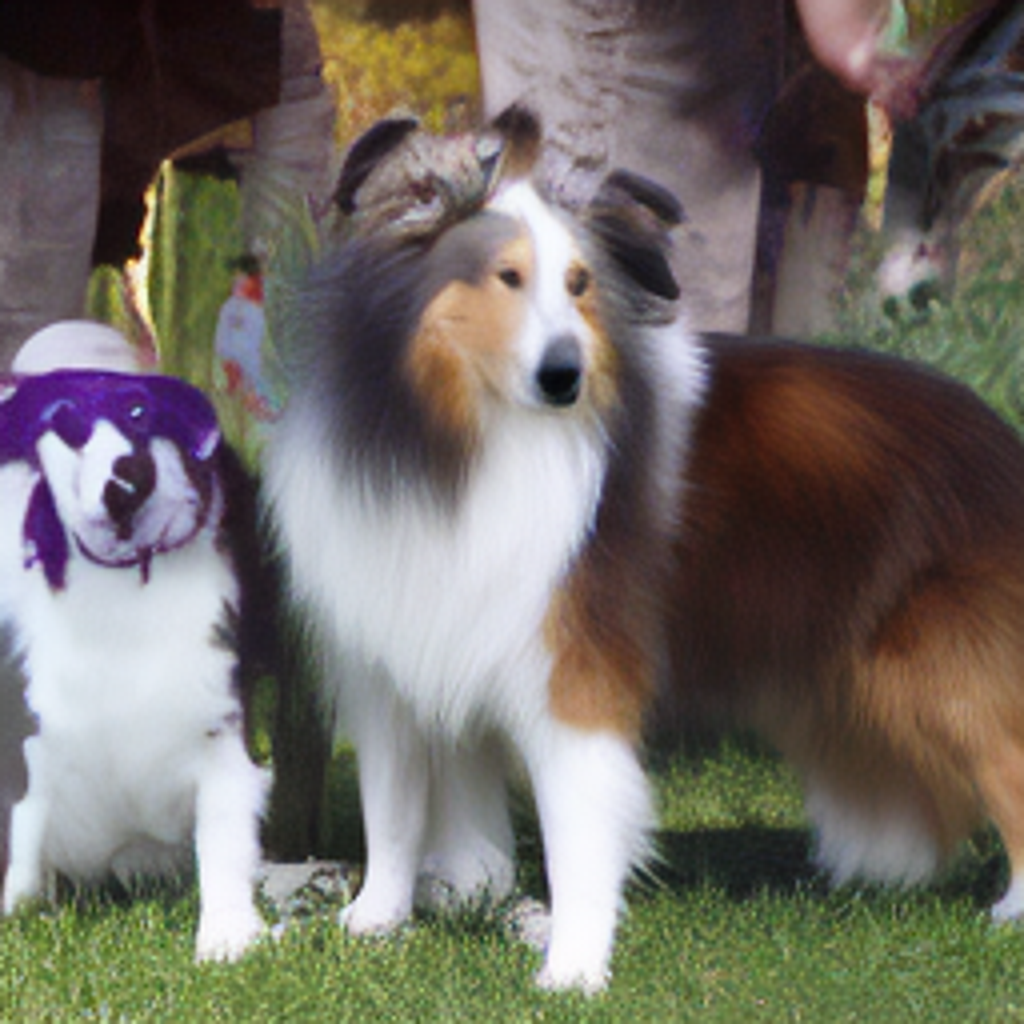} &%
        \includegraphics[width=0.25\textwidth]{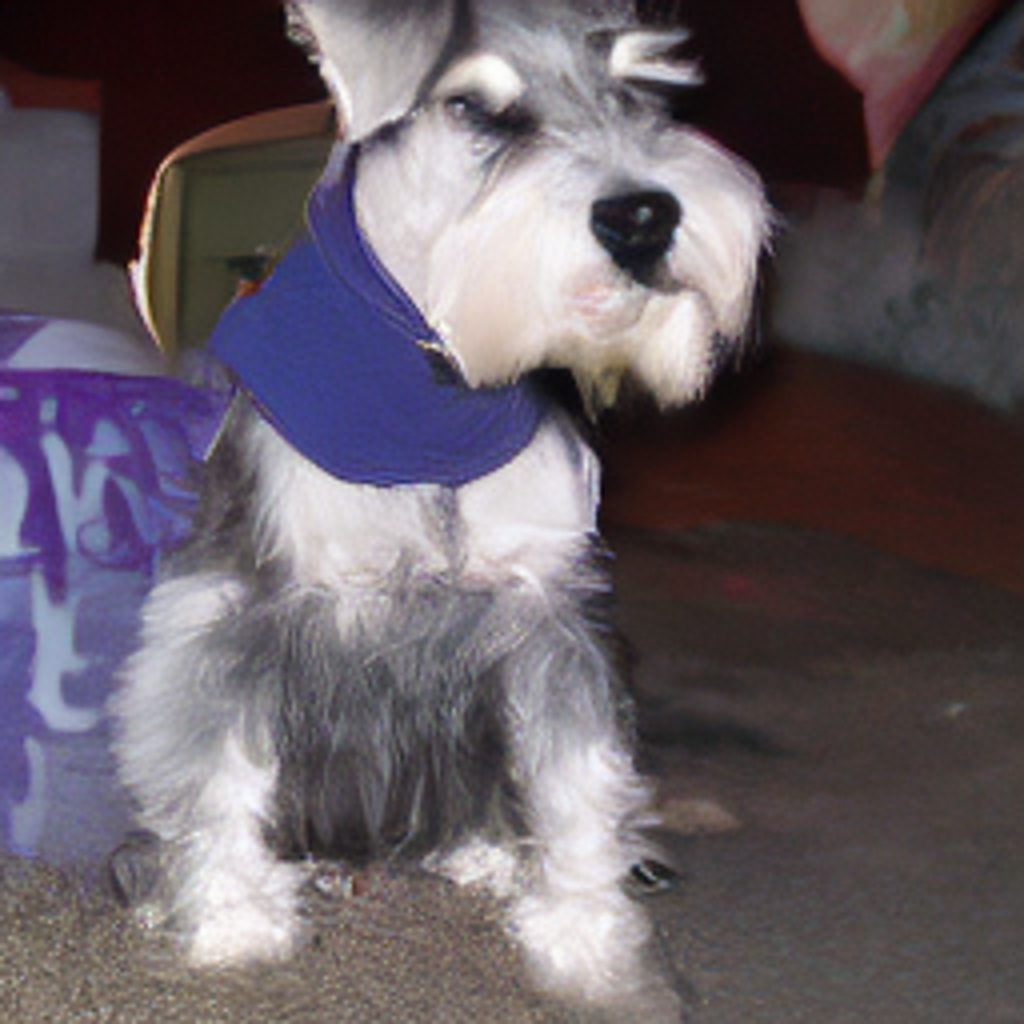} &%
            \includegraphics[width=0.25\textwidth]{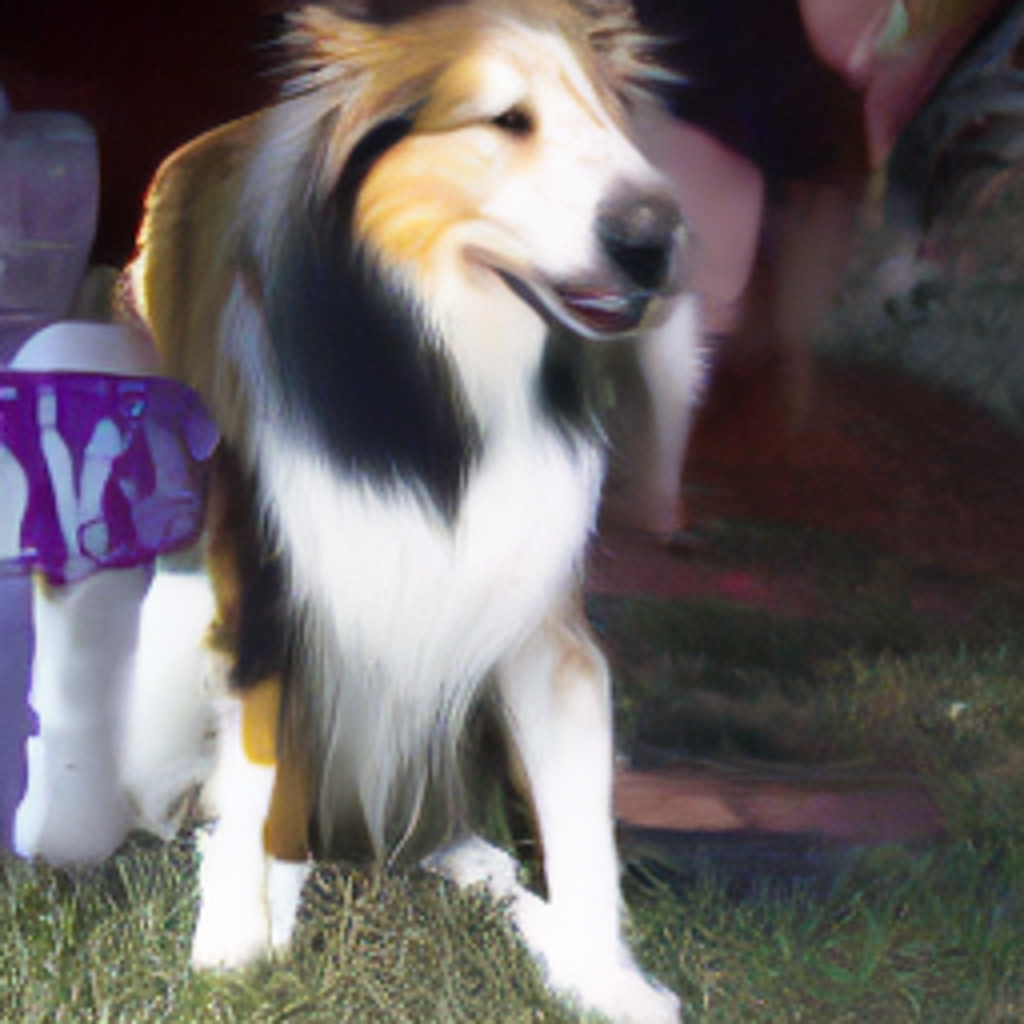} \\%
\includegraphics[width=0.25\textwidth]{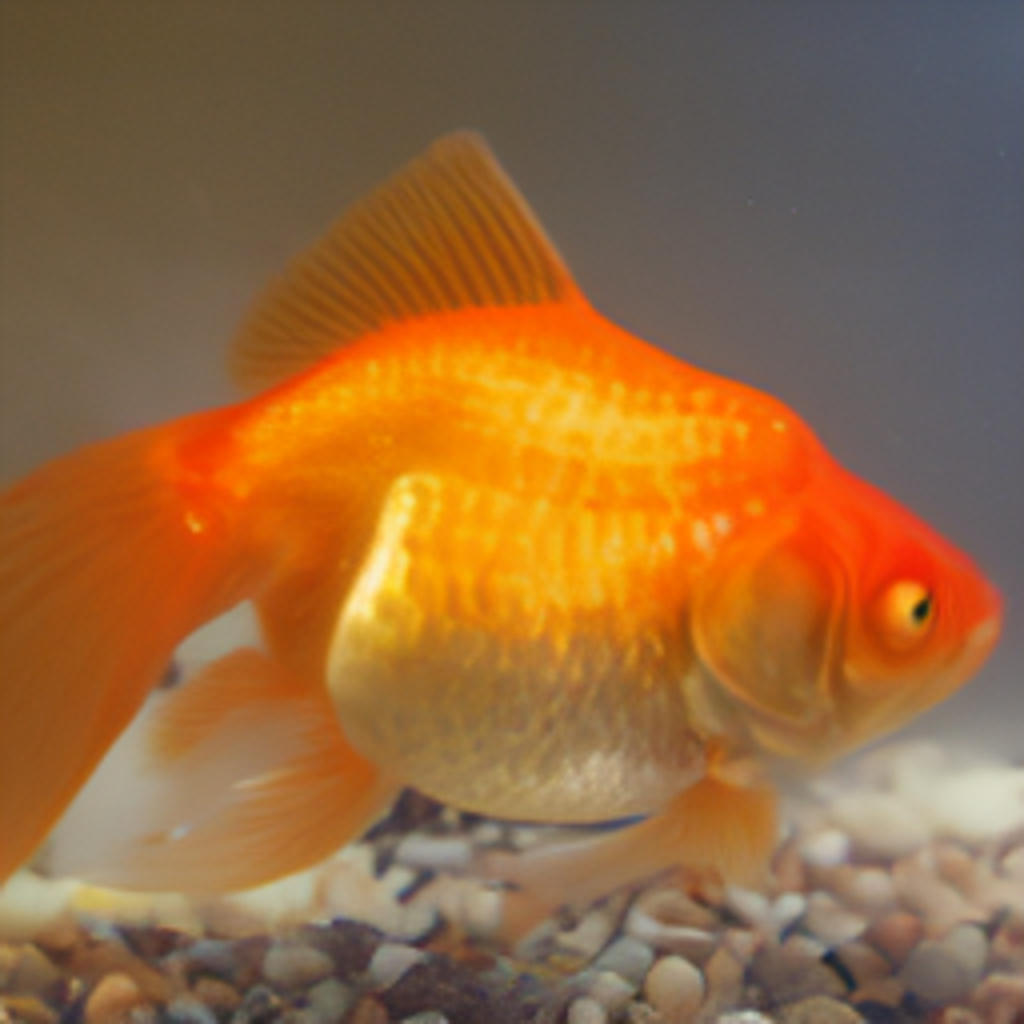} &%
    \includegraphics[width=0.25\textwidth]{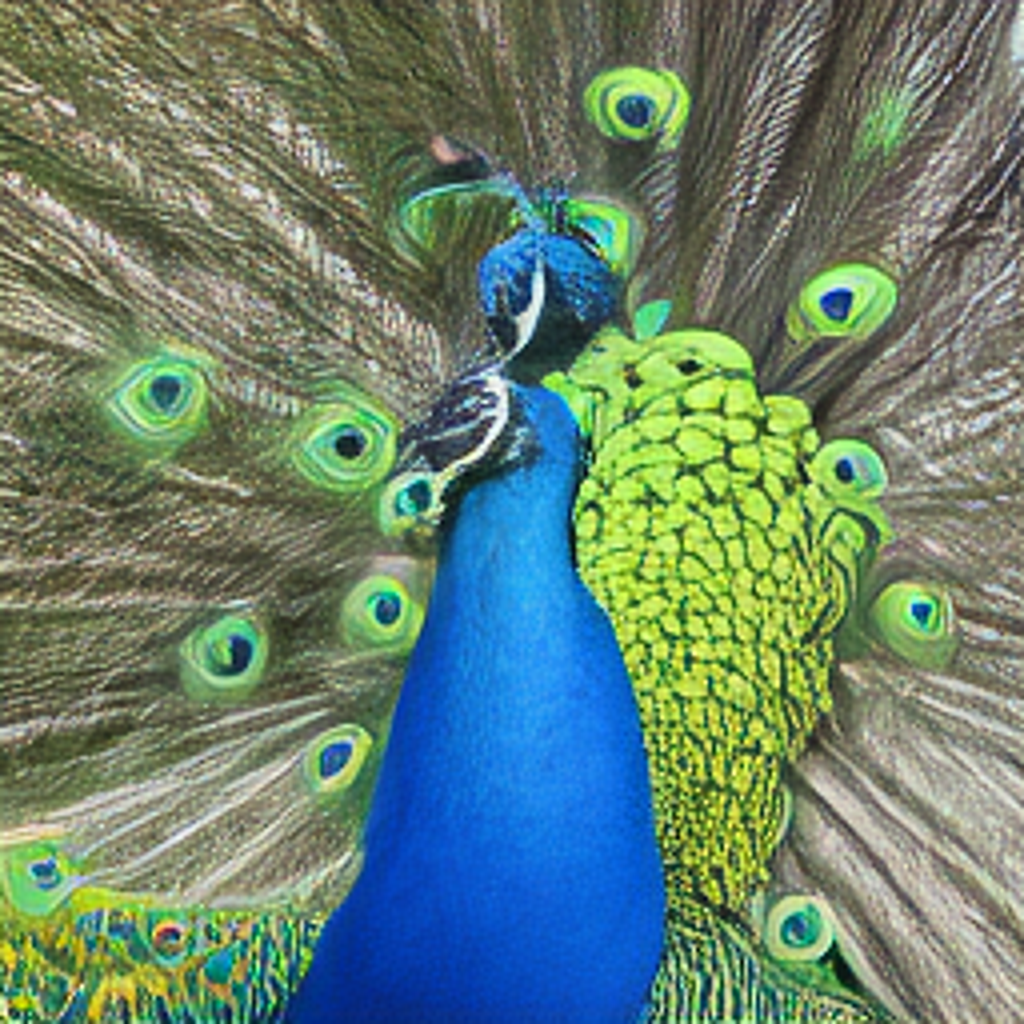} &%
        \includegraphics[width=0.25\textwidth]{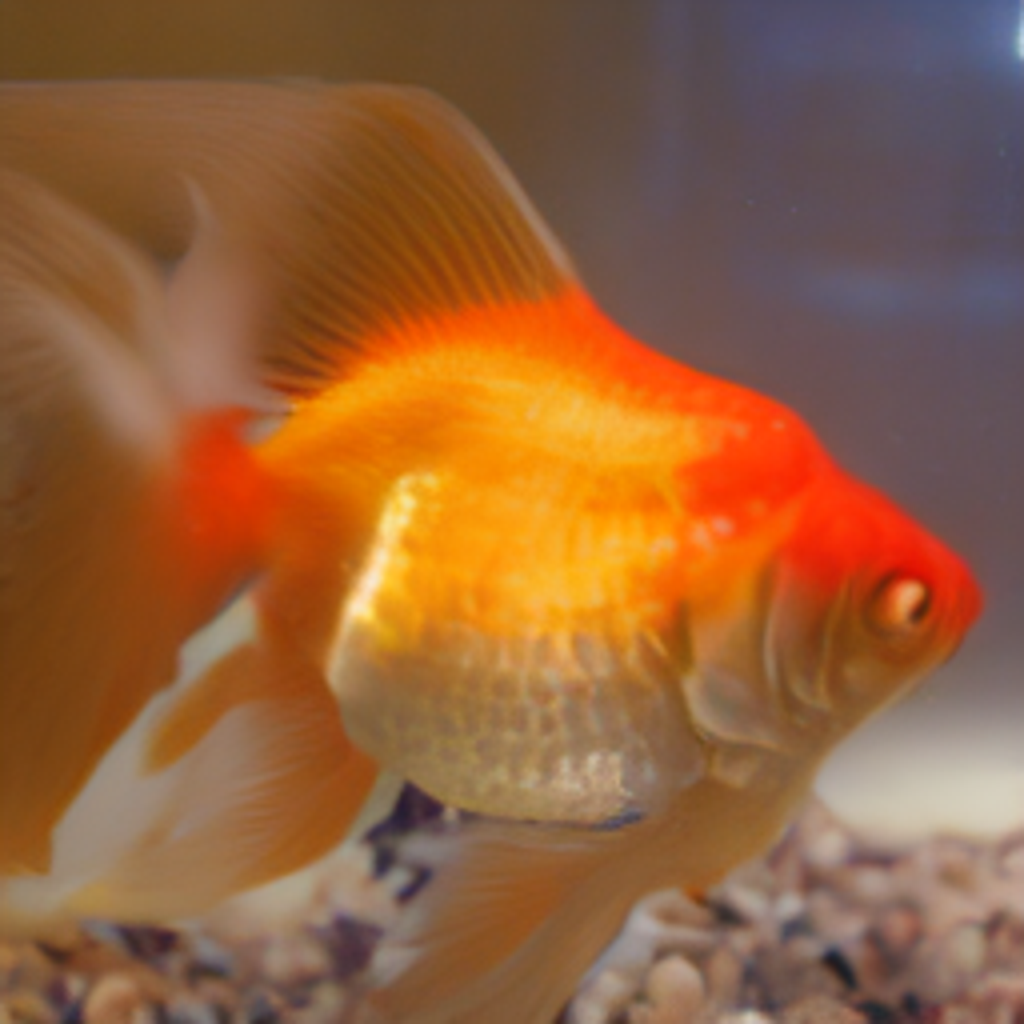} &%
            \includegraphics[width=0.25\textwidth]{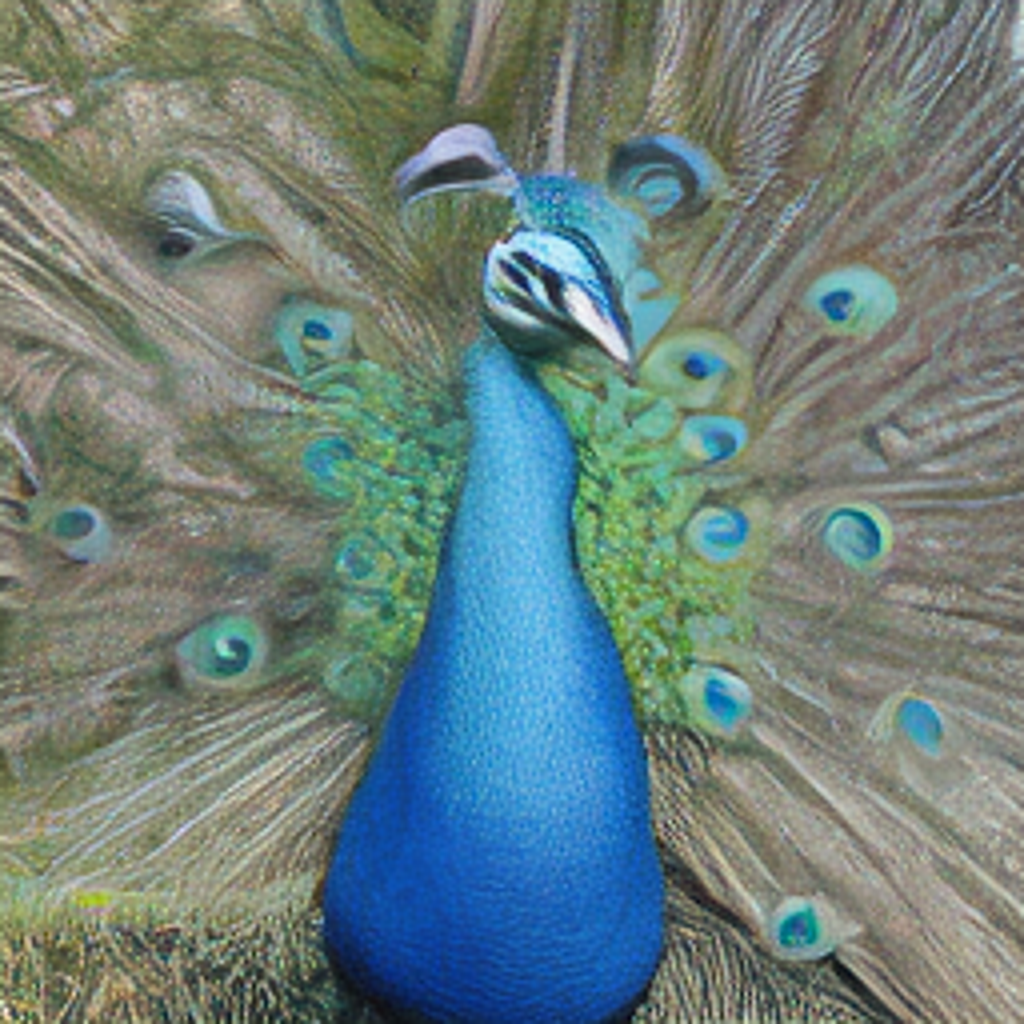} \\%
\includegraphics[width=0.25\textwidth]{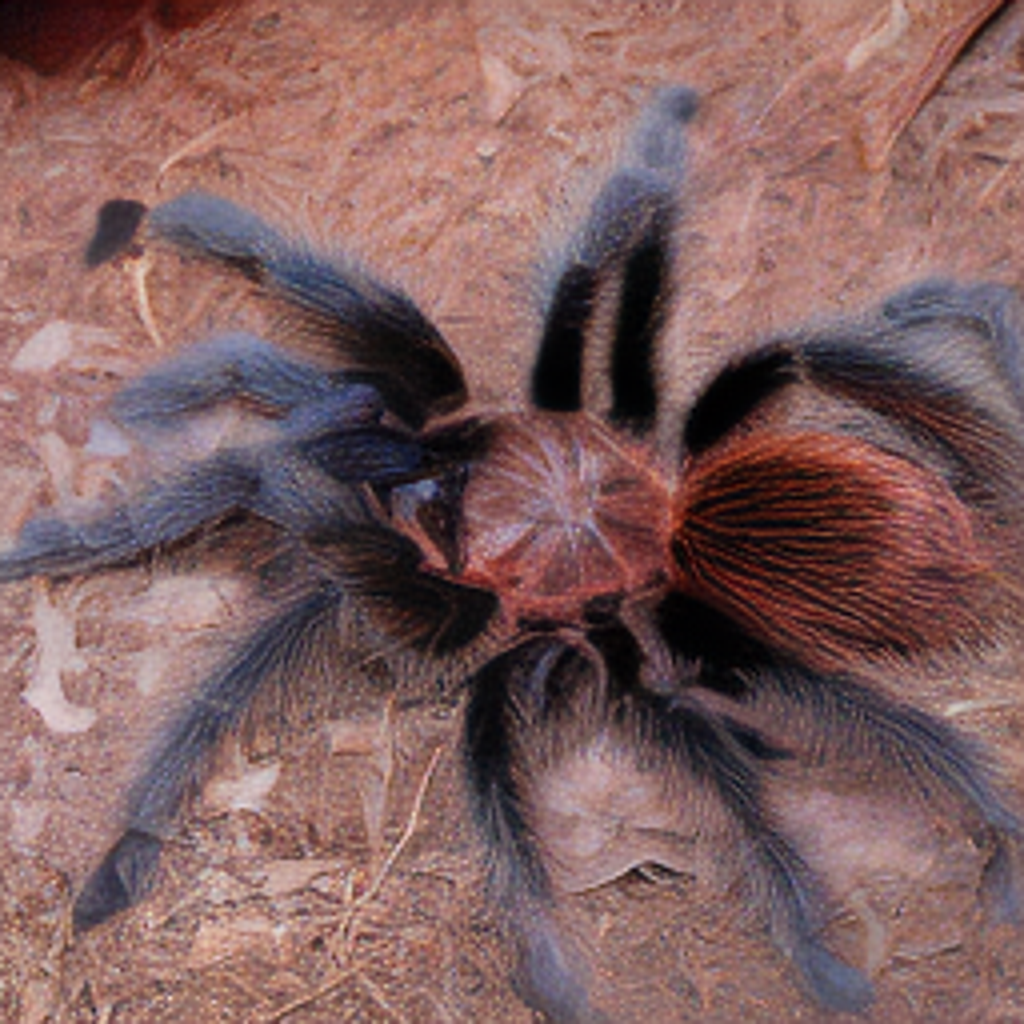} &%
    \includegraphics[width=0.25\textwidth]{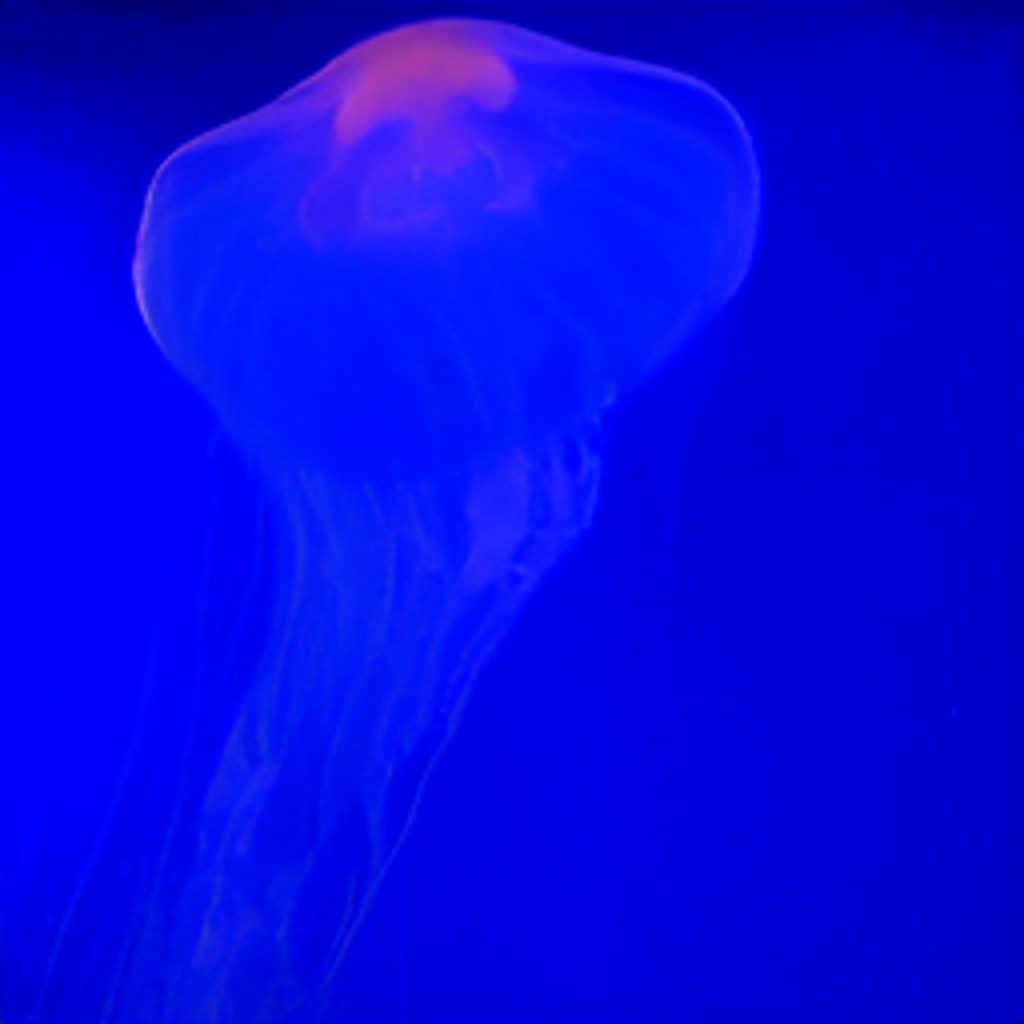} &%
        \includegraphics[width=0.25\textwidth]{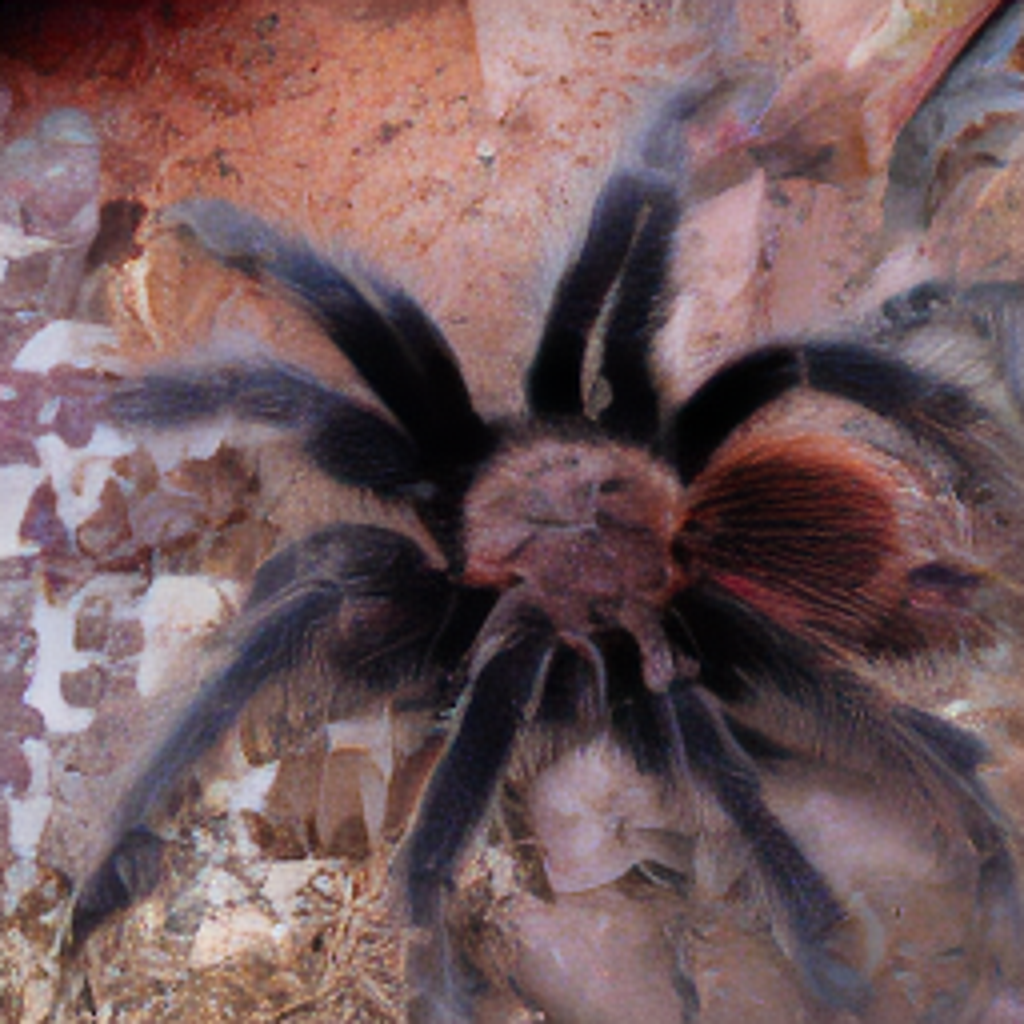} &%
            \includegraphics[width=0.25\textwidth]{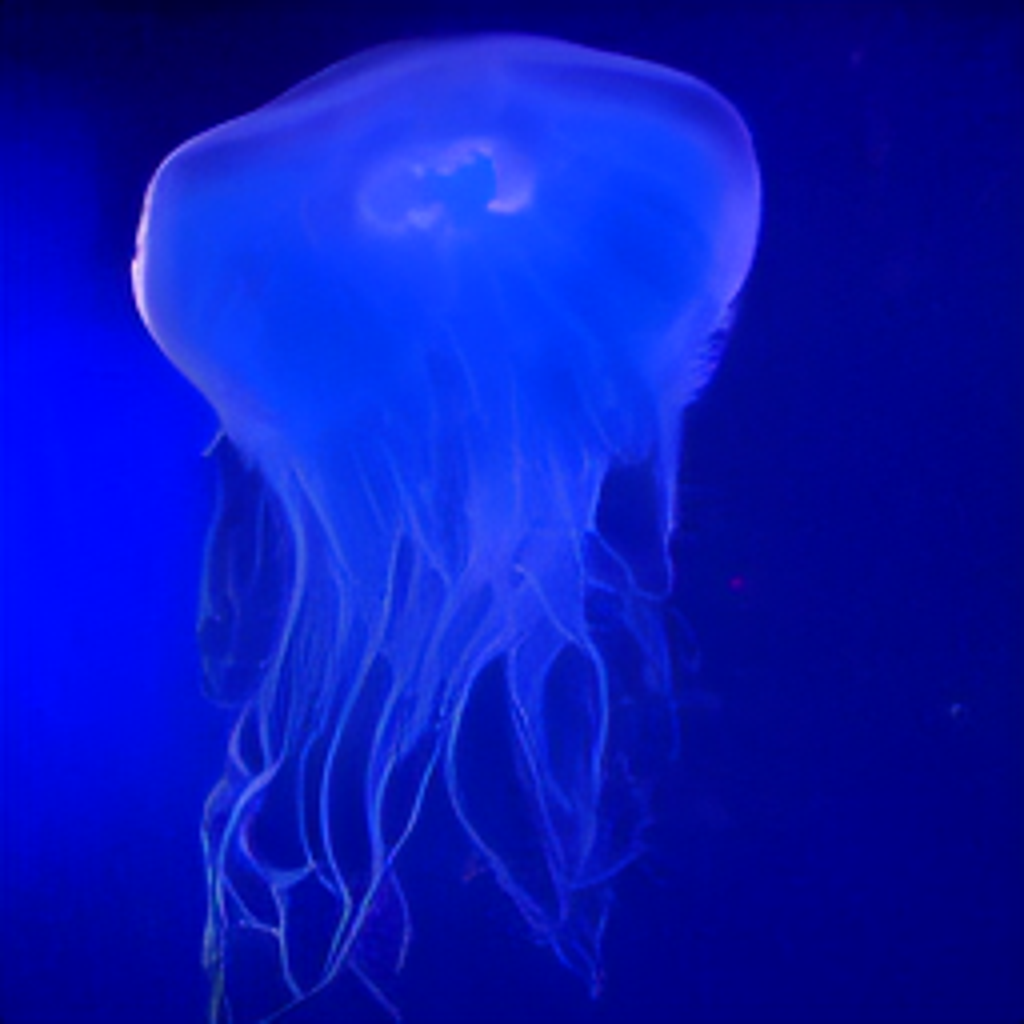} \\%
\bottomrule
\end{tabular}
\end{adjustbox}
\caption{\label{fig:w2a8comparison} \textbf{\emph{Left}}: Sample image generations using the $256\times256$ pixel LDM-4 ImageNet \citep{Rombach2021HighResolutionIS} model with 32-bit floating point weights and activations. \textbf{\emph{Right}}: Images generated after the LDM-4 ImageNet model is quantized to use 2-bit weights and 8-bit activations (W2A8). Only minor loss of quality is apparent. Image generation is conducted under the DDIM sampler \citep{song2020denoising} with 20 sampling steps, a classifier-free guidance scale (CFG) value of 7.5 and a constant random seed of 10.}
\end{figure}

Fortunately, \textbf{model quantization} has emerged as a choice tool for radically shrinking generative models. Quantization methods balance the goal of lossy compression of model weights and/or activations to the maximum extent possible with the desire for minimal loss of generation quality. Many works \citep{shang2023post, li2023q, he2024ptqd, li2024q, so2024temporal, wang2024questlowbitdiffusionmodel, he2024efficientdm, Sui2024BitsFusion1B} have been written on tailoring model quantization methods to the unique challenges posed by diffusion models. EfficientDM \citep{he2024efficientdm} has recently achieved excellent W2A8 (two-bit weights and eight-bit activations) results on the class-conditional LDM-4 ImageNet model \citep{Rombach2021HighResolutionIS} with a compute-efficient post-training quantization (PTQ) inspired approach enabling the calibration of quantized models on consumer hardware. BitsFusion \citep{Sui2024BitsFusion1B} extends the extremely low-bit weight quantization concept to large text-conditional models, achieving excellent quantitative results for generation quality with a 1.99-bit compression of weights for SD 1.5 \citep{Rombach2021HighResolutionIS}. However, their approach involves second-stage quantization-aware training (QAT) on 32 NVIDIA A100 GPUs for 50k iterations, precluding the local production of quantized models by resource-constrained end users.

Despite the many earlier investigations, substantial holes still exist in the model quantization literature on diffusion models. In contrast to the codebook-based Vector Quantization (VQ) approaches such as QUIP\# \citep{tseng2024quip} and AQLM \citep{pmlr-v235-egiazarian24a} that have come to dominate the Pareto frontier of Large Language Model (LLM) quantization, all works on diffusion model quantization to date have focused on Uniform Scalar Quantization (USQ)-based approaches. \textbf{In this paper}, we adapt codebook-based vector quantization to diffusion models, whose heterogenous self-attentive U-Net architecture \citep{RonnebergerFB15, ho2020denoising} and iterative denoising process have no analogue in the NLP domain.

We extensively evaluate our proposed two-stage quantization framework on the standard class-conditional LDM-4 ImageNet \citep{Rombach2021HighResolutionIS}. Our outstanding quantitative and qualitative results show the wide applicability of our low-cost, data-free automated quantization solution, which can efficiently reduce the weights of even large text-conditional models to state-of-the-art sub-2 bit quantization levels with minimal loss of generation quality. We match the weight compression of the best existing works on diffusion model quantization, despite demanding only the quantization-time resources of a single NVIDIA RTX 3090 GPU.

\section{Background and Related Work}
\label{sec:background}

\subsection{Diffusion Models}

Diffusion models \citep{ho2020denoising, song2020denoising} are a class of latent-variable generative model inspired by non-equilibrium thermodynamics, notable for the iterative forward and reverse processes by which they relate the data distribution to an isotropic Gaussian. In the basic case, the forward process is a Markov chain which repeatedly adds Gaussian noise to the sample:
\begin{align} \label{eq:1}
q(\vec{x}_t|\vec{x}_{t-1}) = \mathcal{N}(\vec{x}_{t}; \sqrt{1-\beta_t}\vec{x}_{t-1}, \beta_t\mathbf{I})
\end{align}
where the variance schedule $\beta_t \in (0,1)$ controls the amount of noise added in each of $T$ time steps.
The reverse process is then approximated by a learned conditional distribution:
\begin{align} \label{eq:2}
p_\theta(\vec{x}_{t-1} | \vec{x}_t) &= \mathcal{N}(\vec{x}_{t-1}; \tilde{\vec{\mu}}_{\theta,t}(\vec{x}_t), \tilde{\beta}_t\mathbf{I}).
\end{align}
where at each time-step $\tilde{\vec{\mu}}_{\theta,t}(\vec{x}_t)$ is calculated by a noise estimation network with shared weights. Quantization induces error in the value of $\tilde{\vec{\mu}}_{\theta,t}(\vec{x}_t)$ at each time-step.

The cost of diffusion model inference is determined substantially by the number of time steps at which noise prediction must be carried out as well as the cost of model inference for a single instance of noise prediction. Accelerated sampling strategies such as the DDIM \cite{song2020denoising}, PLMS sampler \cite{liupseudo} and DPM-Solver \cite{lu2022dpm} seek to reduce the number of denoising time steps. In contrast, quantization approaches such as ours instead target the cost of noise prediction for a single denoising step, considered in terms of the amount of GPU video memory (VRAM) utilized as well as the number of computational operations required.

\subsection{Diffusion Model Quantization}
Earlier works on the quantization of diffusion models have included PTQ4DM \citep{shang2023post}, Q-Diffusion \citep{li2023q}, PTQD \cite{he2024ptqd}, Q-DM \citep{li2024q}, TDQ \citep{so2024temporal}, TFMQ-DM \cite{huang2024tfmq}, EfficientDM \cite{he2024efficientdm}, and most recently BitsFusion \cite{Sui2024BitsFusion1B}. These works strike a distinction between quantization of model weights and quantization of model activations; activation quantization is designed to provide benefits on select GPU hardware which provides acceleration for low-bit integer multiplication operations, whereas weight quantization achieves a reduction in the amount of GPU VRAM, system memory or permanent storage capacity used to hold the model weights, even in the absence of specific hardware support for low-bit integer multiplication. BitsFusion \cite{Sui2024BitsFusion1B} entirely eschews activation quantization of Stable Diffusion 1.5 \cite{Rombach2021HighResolutionIS}, carrying out only weight quantization. While we demonstrate compatibility of our approach with activation quantization via our experiments on LDM-4 ImageNet \cite{Rombach2021HighResolutionIS}, activation quantization is likewise not the focus of our work. Instead, we seek to push the boundaries for weight quantization.

\subsection{Quantization Strategies}

Previous works on the quantization of diffusion models such as Q-diffusion \citep{li2023q} have exclusively focused on uniform scalar quantization (USQ), where each weight is individually mapped from its full-precision floating-point representation $w$ to a low-bit integer $\hat{w}$ via a learned affine transformation:
\begin{equation} \label{eq:3}
\hat{w} = \mathrm{s} \cdot \mathrm{clip}(\mathrm{round}(\frac{w}{s}-z), c_\text{min}, c_\text{max}) + z,
\end{equation}
where $c_\text{min}$ and $c_\text{max}$ are the smallest and largest integer representable at the chosen bit-width and $s, z$ are the learnt layer-wise or channel-wise scale factor and zero-point by which the transformation is parameterized. Works such as TDQ \citep{so2024temporal} and TFMQ-DM \cite{huang2024tfmq} have improved the flexibility of USQ by learning separate quantization parameters at each time-step.

Meanwhile, in the parallel field of LLM quantization, recent state-of-the-art works such as QuIP\# \citep{tseng2024quip} and AQLM \citep{pmlr-v235-egiazarian24a, malinovskii2024pv} have achieved impressive results with \emph{Vector Quantization} (VQ) of model weights. Under $k$-bit vector quantization with $M$ codebooks, groups of $d$ weights each are considered as $d$-dimensional vectors $\in \mathbb{R}^d$ jointly replaced with $M$ indices or codes $\in \mathbb{Z}_{kd/M}$ into codebooks $C^{(1)}, \ldots, C^{(M)} \in \mathbb{R}^{2^{kd/M} \times d}$. We seek to extend this approach to diffusion models.

\subsection{Additive quantization} \label{aq}
\textbf{AQLM} \citep{pmlr-v235-egiazarian24a} introduced the use of \emph{Additive Quantization} (AQ) as its vector quantization method, whereby each group of weights is reconstituted as the sum of its indexed codebook vectors according to the following equation:
\begin{equation}
\widehat{\mathbf{W}} {=} \sum_{m=1}^M  C^{(m)}_{b_{1, m}} \oplus \cdots \oplus \sum_{m=1}^M  C^{(m)}_{b_{2^{kg/M}, m}},
\label{eq:4}
\end{equation}
with $\oplus$ as the concatenation operator and $b_{i m} \in \mathbb{R}^{2^{kg/M}}$ as the code assigned to the $i$-th group of weights and $m$-th codebook under $k$-bit quantization, where $g$ is the group size and $M$ the number of codebooks. Quantization in \citet{pmlr-v235-egiazarian24a} is carried out primarily in successive layer-by-layer fashion. The codes and codebooks for the layer are optimized in alternating fashion to minimize $||\mathbf{W} \mathbf{A} - \widehat{\mathbf{W}} \mathbf{A} ||_2^2$ on calibration data, with code optimization carried out via beam search and codebook quantization carried out via Adam \citet{kingma2014adam}. Subsequent Adam optimization of all codebooks simultaneously is suggested as a whole-model PEFT solution. \citet{malinovskii2024pv} instead develop the \emph{PV-Tuning} algorithm for joint optimization of both codes and codebooks against an arbitrary loss on a whole-model basis.

The three important hyperparameters which determine the achieved bit-width under AQLM are the number of codebooks $M$, the group size $g$ and the size of each codebook index in bits, which we may fix as $n=kg/M$ for $k$-bit weight quantization. Regarding contribution to bit-width from the size of the codebook itself, $n=8$ as suggested by \cite{pmlr-v235-egiazarian24a} results in a small codebook of only 256 rows.

In this work, we opt to control only the number of codebooks $M$ as our hyperparameter controlling the amount of lossy compression applied to the weights of a layer. The group size $g$ is set to $g=8$ for fully-connected and $1\times1$ convolutional layers and $g=9$ for $3\times3$ convolutional layers. The latter choice is made so that the weights corresponding to a single input channel and a single output channel correspond to a single group. Furthermore, the codebook indices are set to $n=8$ bits each, so that the size of the codebook itself is small compared to the size of the quantized weight matrix. These choices are comparable to the $2\times8$ scheme proposed in AQLM \cite{pmlr-v235-egiazarian24a}, where for approximately 2-bit quantization of an LLM layer $M=2$ codebooks are used with $g=8$ and $n=8$.
\section{Vector Quantization for Diffusion Models}

Recent works on codebook-based vector quantization of generative models \citep{tseng2024quip, pmlr-v235-egiazarian24a, malinovskii2024pv} have focused on transformer-based LLMs and the quantization of fully-connected or linear layers. Diffusion models differ from LLMs in several key aspects, including the iterative denoising procedure by which they produce a sample and also the U-Net architecture, which features $3\times3$ and $1\times1$ convolutions in addition to linear layers. Ours is the first work to apply vector quantization to diffusion models.

The skeleton of our approach is a two-stage process. In the first stage, we convert each layer of the model to a quantized layer by means of per-layer calibration, so as to minimize a calibration loss
${\arg\min}||\mathbf{W} \mathbf{A} - \widehat{\mathbf{W}} \mathbf{A} ||_2^2$ for each layer independently. In the second stage, we perform knowledge distillation against a full-precision teacher model. This two-stage approach is common for diffusion model quantization and may be recognized in earlier works such as TDQ \cite{huang2024tfmq}, EfficientDM \cite{he2024efficientdm} and BitsFusion \cite{Sui2024BitsFusion1B}. In the following sections, we illustrate the novel modifications we make to this basic structure in order to adapt vector quantization to diffusion models.

\subsection{Stage 1: Layer-By-Layer Calibration}

The current state-of-the-art options for LLM vector quantization are QuIP\# \citep{tseng2024quip}, AQLM \citep{pmlr-v235-egiazarian24a}, and the recently released QTIP \cite{Tseng2024QTIPQW}. QuIP\# and QTIP are both carefully designed to have post-quantization codebooks which barely fit in L1 cache \cite{Tseng2024QTIPQW} for a group size of $g=8$ and codebook indices which are also assumed to be of size $n=8$ bits each in order to enable extremely rapid decompression on GPU, whereas the approach of AQLM is more flexible in terms of quantization hyperparameters. As the weights of $3\times3$ convolutional layers may be naturally considered to occur in groups of size $g=9$ weights, a key design choice of QuIP\# and QTIP would seem to be obviated. Consequently, we choose to use additive vector quantization \cite{pmlr-v235-egiazarian24a} as our VQ method for the first stage of per-layer calibration.

Each layer is quantized independently of other laters, rather than successively as in AQLM \cite{pmlr-v235-egiazarian24a} or Q-Diffusion \cite{li2024q}. This choice is similar to other works which conduct a second stage of knowledge distillation, such as EfficientDm \cite{he2024efficientdm} and BitsFusion \cite{Sui2024BitsFusion1B}. By quantizing layers independently, we make it straightforward to parallelize the quantization process across multiple GPUs.

\subsubsection{Additive Quantization of Convolutional Layers.} \citet{pmlr-v235-egiazarian24a} only describes the AQLM compressed weight format in terms of fully-connected layers, with the inference process represented as the matrix multiplication $\mathbf{Y}=\mathbf{W}\mathbf{X}$. We may however extend it to convolutional layers through applying the sliding window \emph{im2col} transformation common to implementations of convolution \cite{Paszke2019PyTorchAI}. The resulting layer input is the dense matrix $\textbf{X} \in \mathbb{R}^{C_{in} \times k_1 \times k_2, b}$, for a $k_1 \times k_2$ convolution with batch size $b$.

\begin{figure}[htb!]
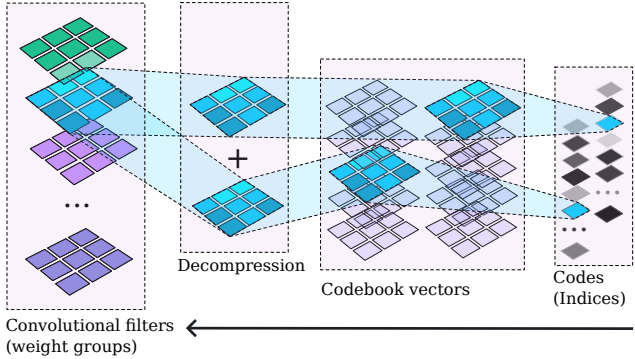

\adjustimage{width=1.0\linewidth,center}{aqlm}  
\caption{\label{fig:aqlm}Additive Quantization (AQ) \citep{pmlr-v235-egiazarian24a}, as applied to a $3\times3$ convolutional kernel with group size $g=9$. The convolutional filter of $3\times3=9$ weights is replaced with an index into each codebook. The indexed codebook vectors are then at inference time summed, producing an approximation to the original weight.}
\end{figure}%

\subsubsection{Convolutional Kernel-Aware Quantization (KAQ)}

We would like to justify our choice of group size $g=9$ for $3\times3$ convolutional layers. We may note that there exists a correlation between those weights corresponding to the same input channel of the convolution. There is also a correlation between those weights corresponding to the same output channel of the convolution. Earlier USQ works on diffusion model quantization \citep{li2023q, huang2024tfmq, so2024temporal, he2024efficientdm, Sui2024BitsFusion1B} have largely chosen to learn $C_{in}$ separate scales $s \in \mathbb{R}^{C_{in}}$ according to $\hat{w} = \mathrm{s} \cdot \mathrm{clip}(\mathrm{round}(\frac{w}{s}-z), c_\text{min}, c_\text{max}) + z$ (Equation \ref{eq:3}).

Ideally, we would like to recognize the correlation of weights according to both output channel and input channel. This is not possible in the case of uniform scalar quantization, as the prohibitive $C_{in} * C_{out}$ number of scale factors required would erase any gains from weight quantization. However, specifically in the case of vector quantization applied to $3\times3$ convolutional kernels, we may still achieve independent quantization of each individual $3\times 3$ filter matrix corresponding to one input and one output channel, via quantizing each such matrix as a vector of $g = 9$ weights.

\begin{figure}[htb!]
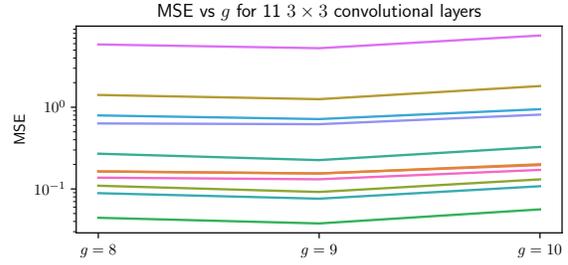

\adjustimage{width=1.0\linewidth,center}{kaq}  
\caption{\label{fig:kaq}The mean square quantization error versus the group size $g$, measured for 11 randomly selected $3\times3$ convolutional layers in the $256 \times 256$ LDM-4 ImageNet \cite{Rombach2021HighResolutionIS} model. Additive quantization à la AQLM \cite{pmlr-v235-egiazarian24a} is performed using a single codebook.}
\end{figure}

Experimentally, this specific choice results in a lower MSE on calibration data for the layer during first-stage quantization compared to both $g=8$ and $g=10$ (Figure \ref{fig:kaq}), despite $g=9$ resulting in approximately 11\% smaller quantized weight matrices.

\subsubsection{Mixed-Precision Strategy (GreedyQuant)}

Previous literature on vector quantization of large generative models has focused on self-attentive large language models, including Llama 2 \citep{Touvron2023Llama2O} and Mistral-7b \citep{Jiang2023Mistral7}. Such models have a relatively homogeneous construction: They are largely composed of self-attention layers and fully-connected multi-layer perceptron blocks. Furthermore, the transformation of each input token to the attention layer into a key, query and value vector as well as the action of the learned projection matrices belonging to each attention head is also typically implemented in terms of the token-wise application of separate fully-connected layers. A "one size fits all layers" approach to quantization is thus enabled, whereby each weight matrix of a fully-connected layer is quantized identically.

\begin{figure}[htbp]
\centering
\begin{minipage}{0.49\linewidth}
\centering
\includegraphics[width=\linewidth]{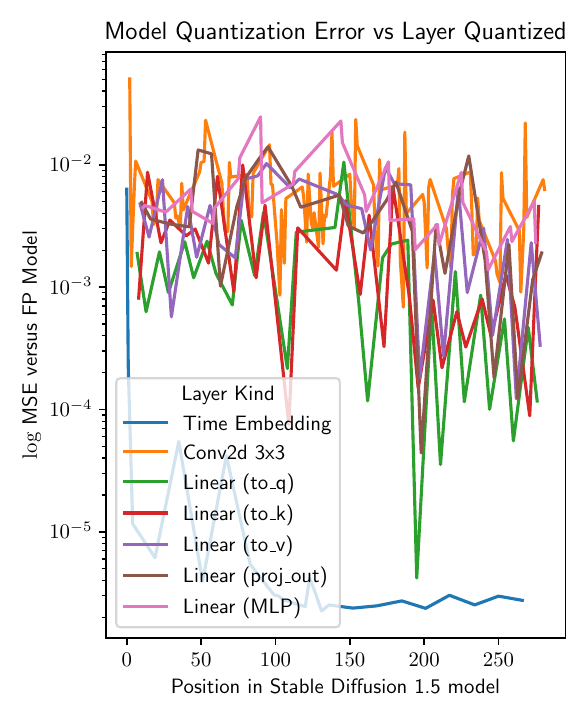}
\end{minipage}
\begin{minipage}{0.49\linewidth}
\centering
\includegraphics[width=\linewidth]{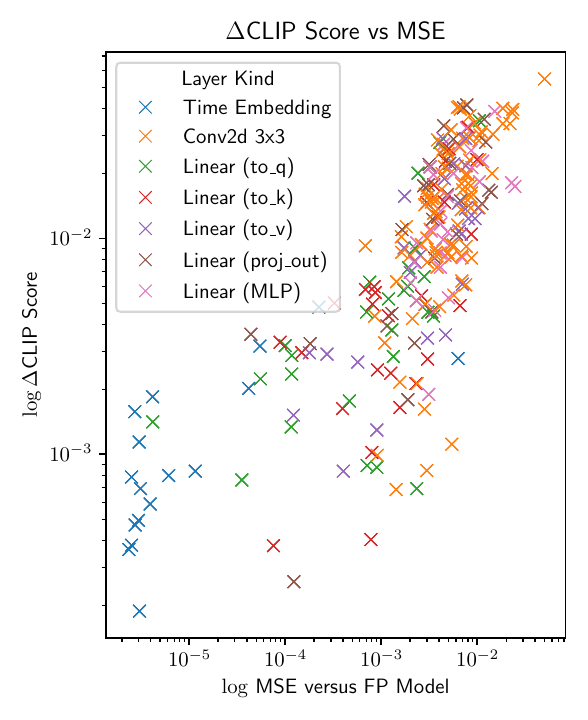}
\end{minipage}
\caption{\textbf{Left}: The mean-square quantization error induced against the output of the full-precision model, as each layer of Stable Diffusion 1.5 
\label{fig:laq}
\cite{Rombach2021HighResolutionIS} is individually quantized using additive quantization \cite{pmlr-v235-egiazarian24a} with one code-book per layer. Colours are used to indicate the kind of layer quantized. The quantization error is not determined by the kind of layer quantized. \textbf{Right}: The reduction in CLIP score versus the induced mean-square quantization error. The two quantities are visibly correlated.}
\end{figure}

In contrast, diffusion models have a highly heterogeneous construction. Latent diffusion models as proposed by \citep{Rombach2021HighResolutionIS} mix self-attention operations with $3\times3$ and $1\times1$ convolutional layers, as well as linear "time embedding" layers used to condition the model on the current denoising timestep. Furthermore, as the receptive field is decreased via downsampling in successive downsampling blocks of the U-Net, the number of channels in each convolutional layer is proportionately increased. Any given choice of quantization hyperparameters thus affects each layer of the U-Net separately, both in terms of the achieved compression factor and the quantization error induced in the outputs of the model. The degree of compression achieved also bears no simple relationship to the error induced in the output of the layer, as illustrated in Figure \ref{fig:laq}. When attempting to quantize the weights of diffusion models to an extremely small overall size, it can thus be useful to determine how much to variably compress each individual layer.

BitsFusion \citep{Sui2024BitsFusion1B} earlier noted the necessity of quantizing different layers of the U-Net to different bit-widths. BitsFusion proposes a complex mixed-precision algorithm requiring considerable manual tuning for this purpose. Each layer $i$ is assigned at each candidate bit-width $b$ a sensitivity score $S_i,_b = M_i,_bN_i^{-\eta}$, where $M_i,_b$ is the MSE induced in the model outputs by quantization of the layer, $N_i$ the parameter count, and $\eta \in [0,1]$ an empirically-derived hyperparameter, which must be re-determined for each model to be quantized. BitsFusion proceeds by quantizing each layer to the minimum $b \in \{1, 2, 3, 4\}$ which meets a chosen sensitivity threshold $S_{i,b} < S_o$, unless the quantization of the layer induces an exceptionally high reduction in CLIP score, in which case it is restored to a higher bit-width.

We note that under our additive quantization approach, the MSE versus the output of the full-precision model and the CLIP score reduction are closely correlated, as shown in Figure \ref{fig:laq}. Furthermore, although the total MSE after many layers are quantized has a roughly additive relationship to the MSE after individual layers are quantized, such a simple relationship does not exist for the CLIP score. We thus choose to calculate for each layer $i$ and for each number of codebooks $M \in \{1, 2, 3, 4\}$ a quantization cost $C_{i,M}=\Delta_{i,M}N_{i,M}^{-1}$, where $\Delta_{i,M}$ denotes the MSE reduction when quantizing layer $i$ using $M$ codebooks, and $N_{i,M}$ the size of the quantized layer in bits. This is the same as the sensitivity score of BitsFusion, with the weighting hyperparamater $\eta$ set to $\eta=1$. The problem of achieving a given total size in bits of the quantized U-Net while minimizing the total quantization cost is then the multiple-choice knapsack problem (MCKP), which we solve via a simple greedy algorithm. We term this approach GreedyQuant.

We verify that a latent diffusion model quantized to on average two bits per weight via GreedyQuant produces superior images to a model of equivalent total size for which all layers are quantized using the same number of codebooks, prior to the second stage of knowledge distillation. However, our ablation shows that this performance advantage is not maintained after knowledge distillation. In fact, after distillation the model for which all layers are quantized using the same number of codebooks, a very simple strategy, outperforms the model quantized using GreedyQuant (Section \ref{ablation}). Thus we elect to not make use of any layer-wise selection of quantization hyperparameters, and instead simply set all layers to be quantized with the same number of codebooks in our final approach. This choice has the advantage that we do not need to attempt quantization of each layer with different numbers of codebooks in order to collect statistics such as the increase in output MSE, thus speeding up our Stage 1 quantization by a factor of roughly four.

\subsection{Stage 2: Knowledge Distillation}
\label{stage2methodology}

In the second stage of our approach, we perform knowledge distillation along the lines of EfficientDM \cite{he2024efficientdm} and BitsFusion \cite{Sui2024BitsFusion1B} in order to directly reduce the overall divergence between the quantized model and the full-precision model. While all of these works in addition to TDQ \cite{so2024temporal} have a second stage of fine-tuning in the quantization pipeline subsequent to the layer-by-layer calibration, they differ substantially in how exactly the fine-tuning is executed. TDQ \cite{so2024temporal} performs quantization-aware retraining of the diffusion model without reference to a feature loss. EfficientDM \cite{he2024efficientdm} and BitsFusion \cite{Abramson2024} both perform knowledge distillation, with the quantized model as student and the full-precision model as teacher. Whereas EfficientDM \cite{he2024efficientdm} seeks to have an efficient second-stage quantization process which may be run within reasonable time on standard consumer GPUs such as the NVIDIA RTX3090, BitsFusion's \cite{Sui2024BitsFusion1B} training process requires the usage of 32 A100 GPUs with a total batch size of 1024 for 50k training iterations. BitsFusion \cite{Sui2024BitsFusion1B} is the only published work to provide quantitative results on compression of a text-conditional model (Stable Diffusion 1.5 \cite{Rombach2021HighResolutionIS}). However, the intense resource demands put the local production of quantized models according to the BitsFusion \cite{Sui2024BitsFusion1B} methodology out of reach for resource-constrained users. As it is common for users to produce, distribute and consume third-party fine-tunes of open-weight diffusion models such as Stable Diffusion \cite{Rombach2021HighResolutionIS}, it is of high interest to develop a quantization method applicable to text-to-image diffusion models which requires only a single consumer-grade GPU to carry out.

The knowledge distillation process proceeds as follows. The full-precision model is used to repeatedly generate batches of sample images from noise, each for a total of $T$ time-steps. The inputs to the full-precision model, constituting partially-noised images from various points in the denoising process, are saved to disk as calibration-set denoising trajectories. Subsequently, in each training step of knowledge distillation, a batch of training inputs is drawn from the saved denoising trajectories. Noise prediction is conducted via both the full-precision teacher and the quantized student. The teacher-student loss
\begin{equation}
L_{\theta,t} = \left\| \boldsymbol{\mu}_\theta (\mathbf{x_t}, t) - \hat{\boldsymbol{\mu}}_\theta (\mathbf{x_t}, t) \right\|_2^2,
\end{equation}
where $\boldsymbol{\mu}_\theta (\mathbf{x_t}, t)$ is the full-precision model and $\hat{\boldsymbol{\mu}}_\theta (\mathbf{x_t}, t)$ the quantized model, is subsequently optimized. 

An important question is in exactly what manner the batches of training inputs are drawn from the calibration-set denoising trajectories. EfficientDM \cite{he2024efficientdm} chooses to conduct \textbf{trajectory-aware sampling}, via drawing each calibration-set trajectory of $T$ sampling steps as a set of $T$ training batches in exactly the same order as denoising was originally conducted. Thus each epoch of $T$ training steps corresponds exactly to the denoising of a single batch of images from random noise for $T$ time-steps. This approach does not strictly require saving of model inputs to disk. BitsFusion \cite{Sui2024BitsFusion1B}, meanwhile, chooses to randomly draw uncorrelated calibration-set samples corresponding to distinct denoising trajectories and time-steps for each training batch (\textbf{random uncorrelated sampling}). However, against a baseline of uniform randomness, BitsFusion \cite{Sui2024BitsFusion1B} finds that it is better to increase the likelihood of sampling earlier time-steps and thus less heavily noised input batches, which induce a higher training loss on average. Furthermore, BitsFusion \cite{Sui2024BitsFusion1B} augments the standard teacher-student loss on the model output with an additional feature loss relating the intermediate features of the teacher and the student.

Ultimately, it is not clear a priori which of these design decisions made by earlier approaches are preferable. Thus we choose to conduct an extensive investigation.

\subsubsection{PV-Tuning}

Many existing solutions for extremely low bit-width vector quantization of large language models, such as QuIP\# \citep{tseng2024quip} and AQLM \citep{pmlr-v235-egiazarian24a} perform only layer-wise independent calibration. This is analogous to our first stage of quantization. AQLM \citep{pmlr-v235-egiazarian24a} suggests that users may use AdamW to optimize the learnt codebook after first-stage quantization. However, AdamW cannot modify the assignment of codebook indices to groups of weights in the quantized weight matrix. 

PV-Tuning \cite{malinovskii2024pv} enables optimizing both the learned codebooks (continuous optimization) and the assignment of codes to groups of weights (discrete optimization) according to an arbitrary loss function. Empirically, we note faster and more stable training with PV-Tuning.

\subsubsection{Trajectory-Aware Sampling}

\begin{figure}[htb!]
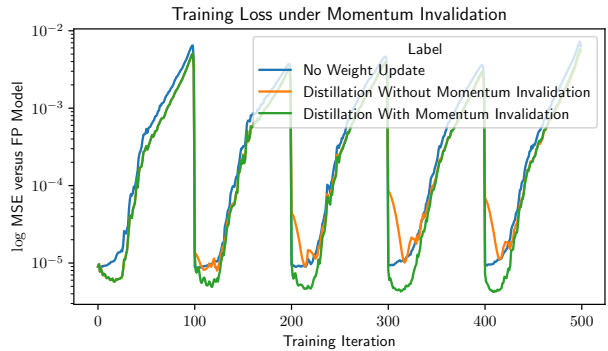

\adjustimage{width=1.0\linewidth,center}{semipv}  
\caption{\label{fig:semipv} Analysis of how convergence of the distillation loss during Stage 2 of the quantization of Stable Diffusion 1.5 \cite{Rombach2021HighResolutionIS} to 2 bits per weight is affected by our choices regarding the momentum of the optimizer. Blue indicates the distillation loss over five epochs of Stage 2 knowledge distillation if no weight updates are made at all and no optimization is performed. Orange indicates the distillation loss if PV-Tuning \citep{malinovskii2024pv} is used naively. The loss can be seen to diverge, as it rises above the baseline of no weight updates. Green indicates the distillation loss if PV-Tuning is used with zeroing of the accumulated momentum and optimizer states at the end of each epoch. The green loss can be seen to decrease in every epoch, confirming convergence of fine-tuning under SeMI-PV.}
\end{figure}

EfficientDM \cite{he2024efficientdm} chooses to conduct \textbf{trajectory-aware sampling}, via drawing each calibration-set trajectory of $T$ sampling steps as a set of $T$ training batches in exactly the same order as denoising was originally conducted. Consequently, there is a heavy and non-regular correlation between successive fine-tuning steps. Specifically, at the end of one epoch and the beginning of another, a batch of almost-fully denoised images is immediately followed by a batch of pure isotropic Gaussian noise. We find that the accumulated optimizer state becomes invalid (Figure \ref{fig:semipv}.) 

In order to resolve this problem, we choose to zero out the momentum and optimizer states of the optimizer at the end of every fine-tuning epoch. The resulting combination of \textbf{momentum invalidation} and \textbf{trajectory-aware sampling} constitutes our first strategy for drawing batches of training inputs from the saved calibration-set denoising trajectories.

\subsubsection{Random Uncorrelated Sampling}
\label{rand_uncorr}

BitsFusion \cite{Sui2024BitsFusion1B} chooses to randomly draw uncorrelated calibration-set samples corresponding to distinct denoising trajectories and time-steps for each training batch. However, against a baseline of uniform randomness, BitsFusion \cite{Sui2024BitsFusion1B} finds that it is better to increase the likelihood of sampling earlier time-steps and thus less heavily noised input batches, which induce a higher training loss on average. BitsFusion \cite{Sui2024BitsFusion1B} additionally augments the teacher-student loss on the model output with a feature loss relating the intermediate features of the teacher and the student.

\begin{figure}
    \centering
    \includegraphics[width=\linewidth]{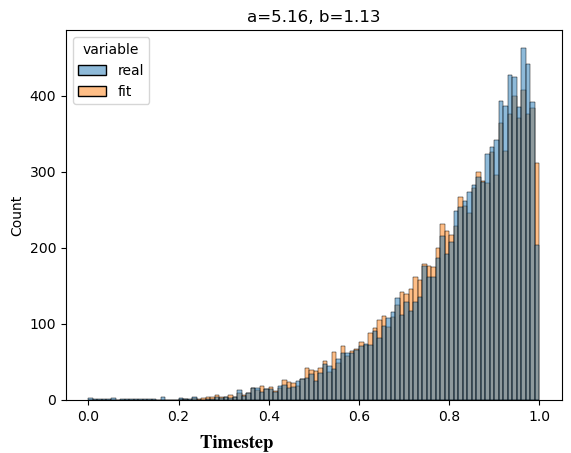}
    \caption{The magnitude of the MSE loss between the output of the teacher and the student model varies greatly with the timestep. We can fit a discrete categorical distribution (shown in blue), or a parametric beta distribution (shown in orange.)}
    \label{fig:timesteps}
\end{figure}

We experiment with both uniform random sampling and weighted sampling. BitsFusion performs weighted sampling via manually fitting a Beta distribution to the average of the training-time loss at each timestep (Figure \ref{fig:timesteps}.) As a simplification of the approach, we choose to simply make use of the empirical distribution of the normalized training loss at each timestep. As in BitsFusion, we make use of an additional feature loss calculated as the mean square error between the intermediate activations of the teacher model and the student model on the training batch, summed over each block of the U-Net, resulting in

\begin{equation}
\begin{aligned}
\mathcal{L}_{\theta,t} &= \mathcal{L}^{\text{Output}}_{\theta, t} + \alpha\, \mathcal{L}^{\text{Feature}}_{\theta, t} \\
&= \left\| \boldsymbol{\mu}_\theta(\mathbf{x}_t, t) - \hat{\boldsymbol{\mu}}_\theta(\mathbf{x}_t, t) \right\|_2^2, \\
&+ \sum_{l=1}^L \left\| \boldsymbol{\mathcal{F}}^l_\theta(\mathbf{x}_t, t) - \hat{\boldsymbol{\mathcal{F}}}^l_\theta(\mathbf{x}_t, t) \right\|_2^2,
\end{aligned}
\end{equation}

where $\boldsymbol{\mathcal{F}}^l_\theta(\mathbf{x}_t, t)$ is the operator retrieving the output features of one of the $L$ up-, down- or mid-blocks in the U-Net of the teacher model and $\alpha$ is a weighting constant compensating for the dissimilar magnitudes of the two terms. In our experiments, the value of $\alpha$ is chosen empirically so that the two loss terms have similar magnitudes.

As an alternative to weighted sampling and keeping in mind the substantial variation in the magnitude of the MSE loss at the output by timestep, we normalize the teacher-student loss across timesteps by dividing the contribution to the teacher-student loss of any given training sample in the training batch by the average teacher-student loss associated with the corresponding time-step. This normalization factor is derived automatically via calculating the loss for a small batch of training samples corresponding to each time-step before conducting knowledge distillation, drawn from the saved calibration-set denoising trajectories. We find in our ablation study (Section \ref{ablation}) that the solution of normalizing the teacher-student loss across time-steps performs substantially better than weighted sampling.

\subsubsection{Ablation Study}
\label{ablation}

We ablate our choice of sampling strategy on the LDM-4 ImageNet model \cite{Rombach2021HighResolutionIS} on the task of weight compression to an average 2 bits per weight, without quantization of activations (Table \ref{tab:ablation}.) In the first stage of quantization, we make use of a calibration dataset of 5120 model inputs, generated via uniform sampling at all inference time-steps as in Q-Diffusion \cite{li2024q}. For the second stage of quantization (knowledge distillation), we first save the denoising trajectories corresponding to the generation of 1280 images sampled uniformly from all image classes over 100 successive denoising steps using the DDIM sampler \cite{song2020denoising} to disk, and subsequently use the PV-Tuning optimizer \cite{malinovskii2024pv} to conduct knowledge distillation for 32,000 iterations with a batch size of 4 and a continuous optimization learning rate of $4e-5$ decaying linearly to $1e-6$ and a discrete optimization learning rate of $1e-4$. Evaluation over 50,000 generated images is subsequently conducted using a DDIM sampler \cite{song2020denoising} with 20 denoising steps and classifier-free guidance scale (CFG) of $7.5$. The reference implementation of the ADM evaluation suite \cite{Dhariwal2021DiffusionMB} to calculate metrics.

\begin{table}[thbp]
\adjustbox{max width=\linewidth}{
\centering
\begin{tabular}{ccccccc}
\hline
\textbf{Method}      & \begin{tabular}[c]{@{}c@{}}\textbf{Bit-width}\\ (W/A)\end{tabular} & \textbf{IS$\uparrow$}     & \textbf{FID$\downarrow$}   & \textbf{sFID$\downarrow$} \\ \hline
    \begin{tabular}[c]{@{}c@{}}Trajectory-Aware Sampling\\+Momentum Invalidation\end{tabular} & 2/32 & {206.78}  & {6.85} & {7.85} \\ \hline
    \begin{tabular}[c]{@{}c@{}}Random Uncorrelated Sampling\end{tabular} & 2/32 & {100.70}  & {17.06} & {12.74} \\
    \begin{tabular}[c]{@{}c@{}}+Weighted Sampling\end{tabular} & 2/32 & {84.03}  & {21.28} & {11.68} \\
    \begin{tabular}[c]{@{}c@{}}-Weighted Sampling\\+Feature Loss\end{tabular} & 2/32 & {241.12}  & {6.75} & {8.79} \\
    \begin{tabular}[c]{@{}c@{}}+Normalised Loss\end{tabular} & 2/32 & \textbf{242.89}  & \textbf{6.23} & \textbf{7.52} \\
    \begin{tabular}[c]{@{}c@{}}+GreedyQuant\end{tabular} & 2/32 & {200.90}  & {6.34} & {6.64} \\ \hline
\end{tabular}}
\caption{Ablation of the components of our method on the class-conditional model LDM-4 ImageNet $256\times256$ \citep{Rombach2021HighResolutionIS}. 50,000 images are generated in total, 50 for each class the model was trained on. Inception Score (IS) \citep{salimans2022progressive}, Fréchet Inception Distance (FID) \citep{Heusel2017GANsTB}, and Spatial FID (sFID) \citep{Salimans2016ImprovedTF} and Precision \citep{Sajjadi2018AssessingGM} calculated using the reference implementation and pre-computed whole-dataset statistics of the ADM evaluation suite 
\citep{Dhariwal2021DiffusionMB}. Results not of our quantized model sourced from EfficientDM \citep{he2024efficientdm}. Image generations performed with classifier-free guidance scale (CFG) of 7.5 using the DDIM sampler of \cite{song2020denoising} with 20 sampling steps.}
\label{tab:ablation}
\end{table}

We find in Table \ref{tab:ablation} that the random uncorrelated sampling strategy of \cite{Sui2024BitsFusion1B} with the addition of the additional feature loss term and normalization of the MSE loss at output outperforms the trajectory-aware sampling strategy of EfficientDM \cite{he2024efficientdm}, but that the weighted sampling of different time-steps according to average training loss seriously deteriorates performance, rather than improving it. Furthermore, we ablate the usage of the earlier proposed greedy layer selection strategy (GreedyQuant) during the first stage of quantization versus the quantization of all layers usng the same number of codebooks. We show that the proposed GreedyQuant actually decreases model accuracy after knowledge distillation, despite resulting in superior generated images prior to knowledge distillation. Consequently, in our subsequent experiments we use \textbf{random uncorrelated sampling} (Section \ref{rand_uncorr}) as our strategy for retrieving training batches during second-stage knowledge distillation, with the PV-Tuning optimizer \cite{malinovskii2024pv}, uniform sampling of training samples from all time-steps, the additional feature loss of \cite{Sui2024BitsFusion1B}, and normalization of the teacher-student loss across time-steps.

\section{Experiments}

\subsection{Evaluation Methodology}

In order to demonstrate the general applicability of our methods, we evaluate our proposed technique on the benchmark of class-conditional generation using the LDM-4 model of \citet{Rombach2021HighResolutionIS} on ImageNet $256 \times 256$ \citep{Deng2009ImageNetAL}. The LDM-4 ImageNet class-conditional generation task is chosen due to its wide adoption by many earlier works, such as Q-Diffusion \cite{li2023q}, PTQ4DM \cite{shang2023post}, TDQ \cite{so2024temporal}, TFMQ-DM \cite{huang2024tfmq}, and EfficientDM \cite{he2024efficientdm}.

\subsection{Quantization Details}

\subsubsection{Stage 1: Layer-By-Layer Calibration}

Uniform sampling of model inputs at all inference time steps is performed as in Q-Diffusion \citep{li2023q}, resulting in a calibration dataset of 5120 model inputs. Layer-by-layer weight quantization is subsequently carried out via AQLM \citep{pmlr-v235-egiazarian24a} with early-stopping at a relative error tolerance of 0.01. In line with earlier works \citep{li2023q, huang2024tfmq, so2024temporal, he2024efficientdm, Sui2024BitsFusion1B}, only the U-Net of the latent diffusion model \cite{Rombach2021HighResolutionIS} is quantized. Furthermore, the first and last convolutional layers of U-Nets are not quantized, due to their extremely small share of the parameter count and model FLOPs. In line with BitsFusion \cite{Sui2024BitsFusion1B}, the weights of time embedding modules are entirely deleted and instead their output at each time-step is stored as a look-up table. For $WkA8$ quantization, a target of $k$ bits per weight on average is used with the GreedyQuant mixed-precision strategy. A group size of $g=9$ is used for $3\times3$ convolutional layers and $g=8$ for all other layers. Codebook indices are $n=8$ bits. The number of codebooks per layer is decided by GreedyQuant where applicable.

\subsubsection{Stage 2: Knowledge Distillation}

For the second stage of quantization (knowledge distillation), we first save the denoising trajectories corresponding to the generation of 1280 images sampled uniformly from all image classes over 100 successive denoising steps using the DDIM sampler \cite{song2020denoising} to disk, and subsequently use the PV-Tuning optimizer \cite{malinovskii2024pv} to conduct knowledge distillation for 32,000 iterations with a batch size of 4 and a continuous optimization learning rate of $4e-5$ decaying linearly to $1e-6$ and a discrete optimization learning rate of $1e-4$.  Sampling of training batches is performed according to \textbf{random uncorrelated sampling} as detailed in Section \ref{rand_uncorr}, with the knowledge distillation loss and normalization of the MSE loss at output, with the PV-Tuning optimizer \cite{malinovskii2024pv}, uniform sampling of training samples from all time-steps, the additional feature loss of \cite{Sui2024BitsFusion1B}, and normalization of the teacher-student loss across time-steps.

\subsubsection{Activation Quantization}

This paper focuses on the quantization of weights, not activations. Activation quantization is performed only for the LDM-4 ImageNet \citep{Rombach2021HighResolutionIS} model, in order to compare with earlier works. We quantize activations to eight bits according to the methodology of Q-Diffusion \citep{li2024q}. We use separate scale factors for each time-step as in EfficientDM \citep{he2024efficientdm} and QuEST \citep{wang2024questlowbitdiffusionmodel}, due to the well-attested better performance. 

\subsection{Class-Conditional Generation}

We compare the generation quality of our quantized LDM-4 ImageNet \cite{Rombach2021HighResolutionIS} model with that of previous works using Inception Score (IS) \citep{salimans2022progressive}, Fréchet Inception Distance (FID) \citep{Heusel2017GANsTB}, Spatial FID (sFID) \citep{Salimans2016ImprovedTF} and Precision \citep{Sajjadi2018AssessingGM}. These metrics are chosen due to their use by earlier works \citep{li2023q, huang2024tfmq, so2024temporal, he2024efficientdm, Sui2024BitsFusion1B}. Our results are displayed in Table \ref{tab:imgnetresult}.

\begin{table}[thbp]
\adjustbox{max width=\linewidth}{
\centering
\begin{tabular}{ccccccc}
\hline
\textbf{Method}      & \begin{tabular}[c]{@{}c@{}}\textbf{Bit-width}\\ (W/A)\end{tabular} & \textbf{IS$\uparrow$}     & \textbf{FID$\downarrow$}   & \textbf{sFID$\downarrow$} & \begin{tabular}[c]{@{}c@{}}\textbf{Precision$\uparrow$}\\ (\%)\end{tabular} \\ \hline
FP          & 32/32     & 364.73      & 11.28    & 7.70    & 93.66     \\ \hline
Q-Diffusion & 4/8       & 336.80      & 9.29     & 9.29    & 91.06     \\
PTQD        & 4/8       & 344.72      & \textbf{8.74}   & 7.98    & 91.69         \\
EfficientDM & 4/8       & 353.83  & 9.93 & \textbf7.34    & 93.10         \\
\ourcell AQUATIC-Diff   & \ourcell 4/8   & \ourcell \textbf{358.20}  & \ourcell 9.77 & \ourcell \textbf{5.78}    & \ourcell \textbf{93.65}         \\ \hline
Q-Diffusion & 2/8       & 49.08      & 43.36     & 17.15    & 43.18         \\
PTQD        & 2/8       & 53.36      & 39.37     & 15.14    & 45.89         \\
EfficientDM & 2/8       & 175.03     & 7.60      & 8.12     & 78.90         \\
\ourcell AQUATIC-Diff        & \ourcell 2/8       & \ourcell \textbf{258.16}  & \ourcell \textbf{6.07} & \ourcell \textbf{6.55}    & \ourcell \textbf{87.73}         \\ \hline
\end{tabular}}
\caption{Performance comparison of our method on the class-conditional model LDM-4 ImageNet $256\times256$ \citep{Rombach2021HighResolutionIS}. 50,000 images are generated in total, 50 for each class the model was trained on. Inception Score (IS) \citep{salimans2022progressive}, Fréchet Inception Distance (FID) \citep{Heusel2017GANsTB}, and Spatial FID (sFID) \citep{Salimans2016ImprovedTF} and Precision \citep{Sajjadi2018AssessingGM} calculated using the reference implementation and pre-computed whole-dataset statistics of the ADM evaluation suite 
\citep{Dhariwal2021DiffusionMB}. Results not of our quantized model sourced from EfficientDM \citep{he2024efficientdm}. Image generations performed with classifier-free guidance scale (CFG) of 7.5 using the DDIM sampler of \cite{song2020denoising} with 20 sampling steps. For our method, the indicated W4A8 and W2A8 quantization settings actually correspond to respectively 3.88 and 1.95 bits per weight, averaged across all parameters of the U-Net.}
\label{tab:imgnetresult}
\end{table}

We achieve impressive results across the board. At all levels of quantization we greatly exceed the IS, FID and sFID of the best existing solution for quantization of the class-conditional LDM-4 ImageNet \cite{Rombach2021HighResolutionIS} model, EfficientDM \cite{he2024efficientdm}, including at the lowest W2A8 (2-bit weight, 4-bit activation) level of quantization. At the W4A8 level of quantization, we achieve FID and sFID that respectively outperform the full-precision model by 1.51 and 1.92 points. At the W2A8 level of quantization, our FID is 1.73 points lower, sFID 1.57 points lower and IS 83.13 points higher than the previous solution of EfficientDM \cite{he2024efficientdm}, the previous state-of-the-art for extremely low-bit quantization of the LDM-4 ImageNet $256\times256$ \cite{Rombach2021HighResolutionIS} model. Amazingly, our model quantized to fewer than 2 bits per weight on average (a ~16x compression) still performs better than the original full-precision model according to the metrics of FID and sFID. This result establishes us as the Pareto frontier for this task, since our solution is the best-performing choice for generation quality at both the 4-bit and 2-bit levels of weight compression, thus achieving the optimal trade-off for compression versus maintenance of output quality.

\subsection{FLOPs Reduction on ImageNet 256x256} \label{flops}

Our key focus in our paper is in the reduction of the RAM or VRAM required for storage of the model weights at inference time, at which we exceed all previous solutions. However, we might also want to reduce the FLOPs required for inference. By default, we may simply decompress the weights from their compressed representation (a very rapid operation) prior to the layer operation. This approach incurs no FLOPs advantage from the weight quantization. Alternatively, we may make use of an efficient inference kernel using efficient LUT-base multiplication. Readers may refer to Section~\ref{appendix}) of our appendix for the procedural details of the FLOPs reduction.

Owing to the substantial technical investment involved, we have not implemented the efficient inference kernel in a manner which actually accelerates model inference. This is typical for papers on DM quantization, and works such as \citet{li2023q, he2024efficientdm} also make claims regarding BOPs (Bitwise OPeration) or latency without a demonstrated speed-up. However, our method is distinguished by the lack of assumptions about hardware support for small integer arithmetic. We display our results in Table~\ref{tab:flopsresult}.

\begin{table}[h]
\caption{FLOPs of our method versus the full-precision model on LDM-4 ImageNet $256\times256$. FLOPs measured for a single forward pass of the U-Net on a batch of size 2 using \emph{fvcore} \citep{fvcore}.}
\label{tab:flopsresult}
\centering
\scalebox{0.90}{
\begin{tabular}{cccccccc}
\hline
\textbf{Method}      & \begin{tabular}[c]{@{}c@{}}\textbf{Bit-width}\\ (W/A)\end{tabular} & \begin{tabular}[c]{@{}c@{}}\textbf{FLOPs}\\ (GFLOPs)\end{tabular} \\ \hline
FP          & 32/32     & 208.78 \\ \hline
AQUATIC-Diff + Infer. Kernel        & 2/8       & 189.54 (-9.22\%) \\ \hline
\end{tabular}
}
\end{table}
\section{Conclusion}

In this work, we have introduced codebook-based additive vector quantization to diffusion models for the first time. In order to account for the unique features of diffusion models, such as the convolutional U-Net and the progressive denoising process, we have introduced techniques such as Convolutional Kernel-Aware Quantization (KAQ), Layer Heterogeneity-Aware Quantization (LAQ), and Selective Momentum Invalidation PV-Tuning (SeMI-PV). Our method has achieved state-of-the-art results in extremely low-bit quantization. We have set a new Pareto frontier on the LDM-4 benchmark at 20 inference steps. Additionally, our approach achieves hardware-agnostic FLOPs savings.

\subsection{Limitations and future work.} Although AQUATIC-Diff achieves excellent results on a variety of metrics, we are not as efficient in terms of quantization-time GPU hours compared to some earlier PTQ+PEFT works, with our quantization process taking approximately 36 hours on an RTX3090 GPU, compared to 3 for EfficientDM \cite{he2024efficientdm}. In part, this results from the slowness of the AQLM layer-wise quantization \citep{pmlr-v235-egiazarian24a} and of the PV-Tuning optimizer \citep{malinovskii2024pv}, in comparison to straight-through estimation using Adam \citep{kingma2014adam} as applied in \citet{he2024efficientdm}. In order to address this, work can be invested in the development of faster gradient-based optimization algorithms for additive vector quantization.

\subsection{Acknowledgement.} This work was conducted using the hardware resources of NTU SYmmetric cryptography and machine Learning Lab (SyLLab). We would also like to thank the first author of \citet{wang2024questlowbitdiffusionmodel} for the generous and helpful correspondence.

{
    \small
    \bibliographystyle{ieeenat_fullname}
    \bibliography{main}
}

\appendix
\section{Appendix} \label{appendix}

\subsection{Proof of FLOPs savings via Efficient Inference Kernel}
\label{app:proof}

Consider a convolutional layer with a weight matrix \( F \) consisting of \( C_{out} \) individual filters \( \{ F_i \}_{i=1}^{C_{\text{out}}} \), where each filter has dimensions \( {C_{in}} \times {h_1} \times {w_1} \). The forward pass on an input \( H \) can be described as the channel-wise concatenation:
\begin{equation}
G = \bigotimes\limits_{i=1}^{C_{out}} H * F_i,
\end{equation}
where \( F \in \mathbb{R}^{C_{out} \times C_{in} \times h_1 \times w_1} \), \( F_i \in \mathbb{R}^{C_{in} \times h_1 \times w_1} \), \( H \in \mathbb{R}^{C_{in} \times h \times w} \), \( H * F_i \in \mathbb{R}^{h \times w} \), and \( * \) denotes the convolution operation (non-batched). Note that we have implicitly padded the convolution so as to keep the spatial dimensions the same. We may now apply the classic formula for FLOPs of a non-batched convolution operation:

\begin{equation}
\text{FLOPs} = C_{\text{out}} \times C_{\text{in}} \times h \times w \times h_1 \times w_1 \times 2.
\end{equation}

Now, instead consider the decompression of a weights matrix quantized via AQLM:

\begin{equation}
\widehat{\mathbf{W}} {=} \sum_{m=1}^M  C^{(m)}_{b_{1, m}} \oplus \cdots \oplus \sum_{m=1}^M  C^{(m)}_{b_{N, m}},
\end{equation}

with $\oplus$ as the concatenation operator and $b_{i m} \in \mathbb{Z}_{2^{k}}$ as the code assigned to the $i$-th group of weights and $m$-th codebook under quantization with a $k$-bit codebook, where $k$ is the number of bits used for storing each index into the codebook, $N$ is the number of weight groups $M$ the number of codebooks. We may think instead of the decompression of a convolutional filter where each channel-wise slice of $h_1 \times w_1$ weights is quantized as a vector:

\begin{equation}
F_i {=} \sum_{m=1}^M  C^{(m)}_{b_{1, m}} \oplus \cdots \oplus \sum_{m=1}^M  C^{(m)}_{b_{2^{h_1 w_1}, m}},
\end{equation}

with $\oplus$ as instead the stacking operator, so that the tensor dimensions work out. Substitute:

\begin{equation}
G = \bigotimes\limits_{i=1}^{C_{\text{out}}} H * \left( \sum_{m=1}^M C^{(m)}_{b_{1, m}} \oplus \cdots \oplus \sum_{m=1}^M C^{(m)}_{b_{2^{h_1 w_1}, m}} \right).
\end{equation}

A rearrangement, keeping in mind the manner in which convolution commutes with summation and stacking, grants us:

\begin{equation}
G = \bigotimes\limits_{i=1}^{C_{\text{out}}} \sum_{j=1}^{C_{\text{in}}} \left(\sum_{m=1}^M H_j * C^{(m)}_{b_{1, m}} \oplus \cdots \oplus \sum_{m=1}^M H_j *  C^{(m)}_{b_{2^{h_1 w_1}, m}} \right).
\end{equation}

We may at this point do the tedious work of counting the FLOPs:

\begin{equation}
\begin{split}
\text{Total FLOPs} = & \, M \times 2^k \times C_{in} \times h \times w \times h_1 \times w_1 \; \text{ multiplications} \;  + \\
& \, M \times 2^k \times C_{in} \times h \times w \times (h_1 \times w_1 - 1) \;  \text{ additions} \;  + \\
& \, M \times C_{out} \times C_{in} \times h \times w \;  \text{ additions}.
\end{split}
\end{equation}

Landing us at $\frac{9(2^k-1)M}{2\cdot9-M} \leq C_{out}$ as the breakpoint at which our FLOPs count goes down for a $3\times3$ 2-D convolutional kernel quantized using $M$ codebooks and $k$-bit codebook indices.

\end{document}